\pdfoutput=1

\documentclass[11pt]{article}

\usepackage[]{ACL2023}

\usepackage{times}
\usepackage{latexsym}
\usepackage{graphicx}
\usepackage[T1]{fontenc}

\usepackage[utf8]{inputenc}

\usepackage{microtype}

\usepackage{inconsolata}
\usepackage{graphicx}
\usepackage{subcaption}

\title{Machine Translation with Large Language Models: Decoder Only vs. Encoder-Decoder}

\author{Abhinav P.M.\\
  Calicut University \\
  \And
  SujayKumar Reddy M \\
  VIT University \\\\
  \And
  Dr. Oswald Christopher  \\
  Assistant Professor \\
  National Institute of Technology, Trichy \\
}

\begin{document}
\maketitle
\begin{abstract}

This project, titled "Machine Translation with Large Language Models: Decoder-only vs. Encoder-Decoder," aims to develop a multilingual machine translation (MT) model. Focused on Indian regional languages, especially Telugu, Tamil, and Malayalam, the model seeks to enable accurate and contextually appropriate translations across diverse language pairs. By comparing Decoder-only and Encoder-Decoder architectures, the project aims to optimize translation quality and efficiency, advancing cross-linguistic communication tools.The primary objective is to develop a model capable of delivering high-quality translations that are accurate and contextually appropriate. By leveraging large language models, specifically comparing the effectiveness of Decoder-only and Encoder-Decoder architectures, the project seeks to optimize translation performance and efficiency across multilingual contexts. Through rigorous experimentation and analysis, this project aims to advance the field of machine translation, contributing valuable insights into the effectiveness of different model architectures and paving the way for enhanced cross-linguistic communication tools.

\end{abstract}


\section{Introduction}
Machine Translation (MT) has witnessed significant advancements with the advent of Large Language Models (LLMs), which have revolutionized the field by offering robust capabilities in processing and translating natural language. These models, such as mT5 and LLaMA 2, vary in architecture from decoder-only designs to more complex encoder-decoder frameworks, each tailored to address specific challenges in multilingual translation tasks.
This work delves into the comparative analysis of LLMs across different architectural paradigms: decoder-only (1-1 and 1-many) and encoder-decoder (1-1, many-1, 1-many, and many-many). Our primary objectives are twofold: first, to explore how these models perform in bi-lingual and multilingual language translation scenarios, and second, to evaluate the effectiveness of encoder-decoder transformer models in Neural Machine Translation (NMT) compared to smaller, decoder-only models when trained under similar conditions.
Key considerations include assessing the impact of context length—measured in the number of tokens—on translation quality and efficiency for both architectural setups. By conducting comprehensive experiments and performance evaluations on datasets like FLORES-101 and TED Talks, we aim to provide insights into the optimal use cases and trade-offs associated with different LLM configurations in the realm of MT.

 The code available on GitHub for In-context-learning (ICL), Baseline Model Development, and sample notebooks for finetuning at the following link \footnote{\url{https://github.com/sujaykumarmag/iasnlp}}.

\section{Related Work}
\cite{1} investigated the development of a universal NMT system capable of translating between 103 languages using over 25 billion training examples. The study emphasized the effectiveness of transfer learning for low-resource languages while maintaining high-quality translation for high-resource languages. By exploring complexities such as diverse scripting systems, data imbalance, and model capacity, the research compared multilingual NMT with bilingual baselines. The findings highlighted challenges in scaling models, balancing data distribution, and mitigating domain noise, offering insights for future research in universal machine translation. \cite{2} explored the limits of multilingual NMT by training models to translate between 102 languages and English. Extensive experiments were conducted using the TED Talks multilingual corpus, revealing that massively multilingual models outperform previous state-of-the-art methods in low-resource settings while supporting up to 59 languages. The study analyzed different training setups and highlighted the trade-offs between translation quality and modeling decisions. Results demonstrated that the multilingual models exceeded strong bilingual baselines, indicating promising directions for future research in massively multilingual NMT.

The study by \cite{3} systematically investigated the performance and factors affecting LLMs in multilingual machine translation, finding that models like GPT-4, despite outperforming the strong supervised baseline NLLB in 40.91\% of translation directions, still lag behind commercial systems like Google Translate, particularly for low-resource languages . They used the FLORES-101 dataset to benchmark translation quality and evaluated eight popular LLMs, including ChatGPT and GPT-4 . Their findings also highlighted that cross-lingual exemplars provided better guidance for low-resource translations than same-language exemplars. \cite{4} introduced BERT, a Bidirectional Encoder Representations from Transformers, which pretrains deep bidirectional representations from unlabeled text, conditioning on both left and right context across all layers. This design allows BERT to achieve state-of-the-art results on various natural language processing tasks with minimal task-specific architecture changes. BERT significantly improved performance on tasks such as GLUE, MultiNLI, and SQuAD v1.1, demonstrating absolute improvements in accuracy and F1 scores across different benchmarks \cite{4}. The Transformer architecture introduced by \cite{5} in 2017 revolutionized natural language  processing (NLP) by employing self-attention mechanisms. This discovery transformed  NLP and established a foundation for subsequently developed language translation 
models.

\section{Proposed Methodology and Experimental Results}

The proposed methodology aims to evaluate and compare the performance of Encoder-Decoder and Decoder-only models in natural language processing tasks. This methodology is structured into several key phases, including data preparation, model design, training, and evaluation. Each phase is described in detail below.

\subsection{Incontext Learning}

Our first approach in the Machine Translation is In-Context Learning using Few Shot Learning. A brief description about In-Context Learning and its working is given below.

In-Context Learning allows language models to learn tasks using only a few examples \cite{ICL}. It is often seen as a prompt engineering task for Few-Shot Learning. In this method, Machine Translation pairs (\(<X>=<Y>\)), where \(X\) is the source sentence and \(Y\) is the target sentence, are provided using a template \(T\) \cite{zhu2023multilingual}. The In-Context Exemplars, which include \(<X>=<Y>\) pairs, serve as a strong recipe for generating the best outputs from the model, as noted by Wu et al. \cite{wu2023openicl}.

A prompt \(P\) is defined as \(T(X_{1}, Y_{1}) \oplus T(X_{2}, Y_{2}) \oplus \cdots \oplus T(X_{n}, Y_{n})\), where \(\oplus\) refers to concatenation, and \(n\) represents the number of samples \cite{zhu2023multilingual}.

Our approach to In-Context Learning utilizes 3-shot learning, where the prompt to the model is structured as illustrated in Figure 1.

\begin{figure}
    \centering
    \includegraphics[scale=0.35]{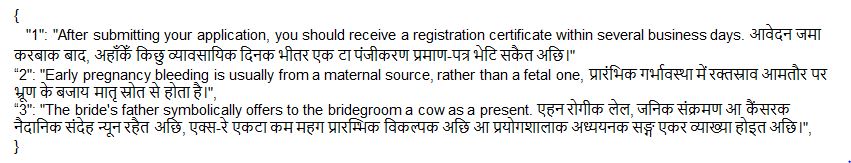}
    \caption{A sample prompt to the In-Context Learning}
    \label{fig:icl}
\end{figure}

For the evaluation of the In-context Learning, we used Hindi, Malayalam, Telugu, Tamil and Marathi languages. A 3-short learning using XGLM and mT5 is used with BLEU metric. 

\subsubsection{Dataset Used}

We used BPCC Wiki MT Dataset\cite{BPCC} which had 16k-50k translation samples. It has English to 22 other Indian Language Pairs with a context length of each sentence pair being 40-200 characters long. 

\subsubsection{ICL Experimental Results}
The Architectures that we used are 1. Decoder Only - XGLM.  The reason for XGLM is it generates moderate translation with 500 million parameters and builds bi-lingual mapping between non-English and English \cite{zhu2023multilingual}. 2. Encoder-Decoder - mT5 because of its capability for multilingual translation (mT5-base) and contains 300 million parameters. A sample reference and its prediction in mT5 and XGLM is given in Figure 2. The experimental results of In-Context Learning for Decoder-only and Encoder-Decoder Architectures are given in Figure 3. 

\begin{figure}
    \centering
    \includegraphics[scale=0.45]{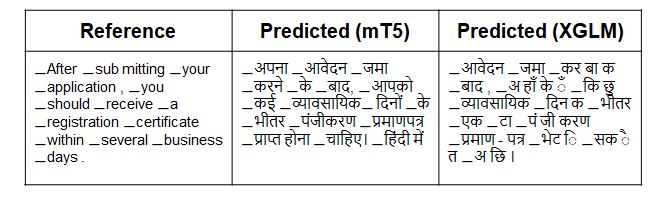}
    \caption{A sample reference results for the Predicted XGLM and mT5}
    \label{fig:icl}
\end{figure}

\begin{figure}
    \centering
    \includegraphics[scale=0.5]{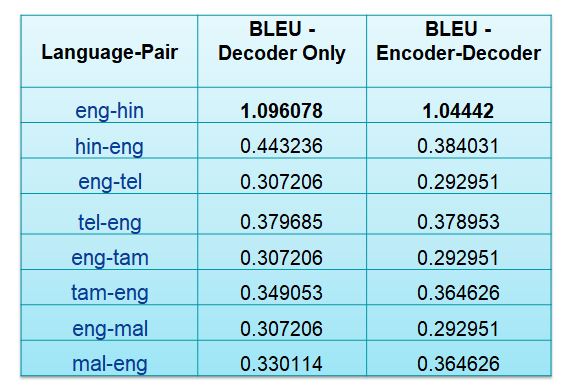}
    \caption{Experimental Results of In-Context Learning - Language Pairs and their BLEU Scores}
    \label{fig:icl}
\end{figure}

\subsection{Finetuning}

\begin{figure}
    \centering
    \includegraphics[scale=0.3]{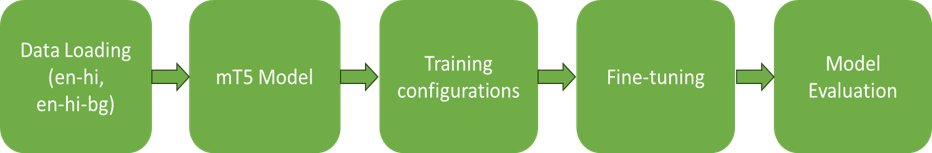}
    \caption{Workflow of mT5 Fine-tuning (Decoder only model)}
    \label{fig:mT5}
\end{figure}

\begin{figure}
    \centering
    \includegraphics[scale=0.3]{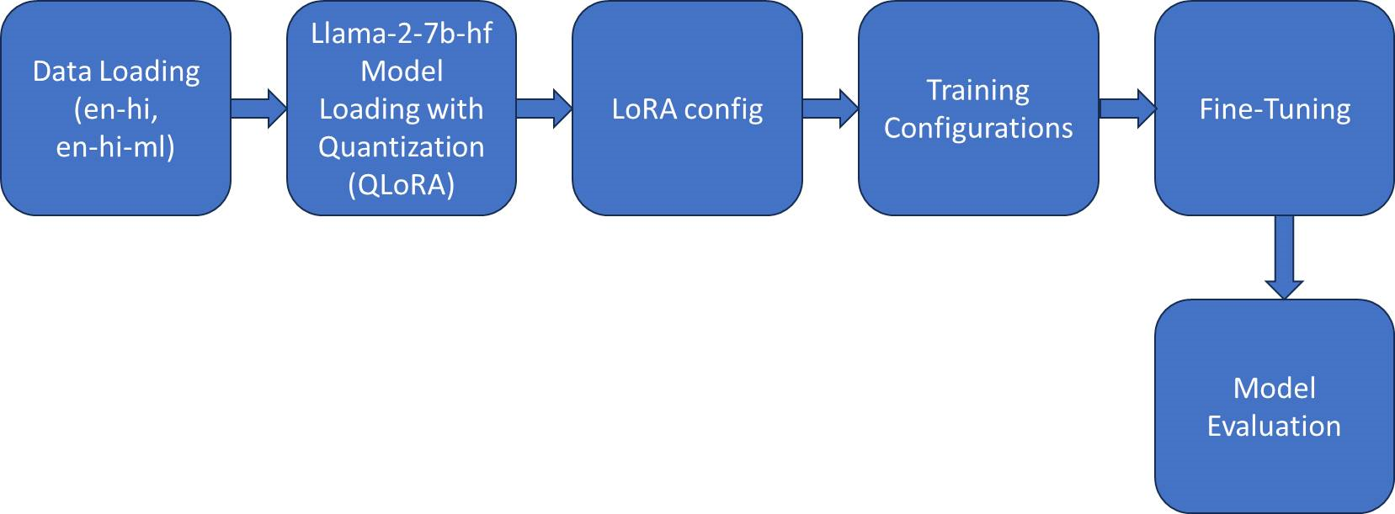}
    \caption{Workflow of Llama2 Fine-tuning (Decoder only model)}
    \label{fig:llam2}
\end{figure}
In this section, we describe the fine-tuning processes of the mT5 and LLaMA 2 models for multilingual and bilingual machine translation tasks. These models were selected because of their superior performance in natural language processing tasks. We meticulously detailed the workflow and evaluation metrics used to gauge their effectiveness.
We began with the mT5 model, an encoder-decoder model fine-tuned for translation tasks involving English to Hindi (en-hi) and English to Hindi-Bengali (en-hi-bg) pairs. The process started with data loading, where we prepared datasets for the specified language pairs. The mT5 model was selected for its robust multilingual capability. We configured the training parameters, including hyperparameters, learning rate, batch size, and optimization algorithms, to ensure an optimal setup for fine-tuning. The model was fine-tuned on the datasets, allowing it to adapt to the translation tasks. The evaluation metrics used for the fine-tuning task were BLEU, chrF, and TER.

Next, we focused on the LLaMA 2 model, a decoder-only model fine-tuned for English-to-Hindi (en-hi) and English-to-Hindi-Malayalam (en-hi-ml) translation tasks. The process began with data loading for the specified language pairs. The LLaMA-2-7b-hf model was loaded with quantization (QLoRA) to enhance the performance and reduce computational demands. Quantization with QLoRA helps in reducing the model size and computational requirements without significantly compromising the model's performance. We set up LoRA configurations to facilitate efficient fine-tuning, followed by establishing training configurations, including hyperparameters and optimization settings. The model was then fine-tuned on the datasets with evaluation metrics, such as BLEU scores, used to assess performance. The LLaMA 2 model exhibited superior performance in one-to-one (1-1) translation tasks compared with one-to-many (1-many) tasks. Visualization of the loss over time indicated a steady decrease, confirming effective fine-tuning for both the bilingual and multilingual mT5 models. To feed data into the models, we used specific prompt templates: source text \#hi\#> target text for English-to-Hindi translations and source text \#ml\#> target text for English-to-Malayalam translations, where \#hi\#> and \#ml\#> signify the target language.

\subsubsection{Finetuning Experimental Results}

\begin{figure}[ht]
    \centering
    \begin{subfigure}[b]{0.4\textwidth}
        \centering
        \includegraphics[width=\textwidth]{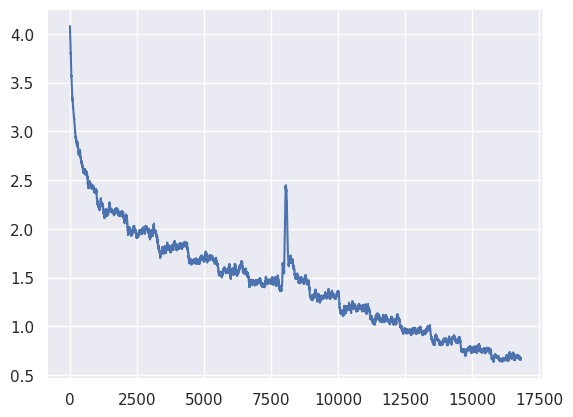}
        \caption{mT5 Multilingual eng, hi, bg (all pairs)}
        \label{fig:mT5-multi}
    \end{subfigure}
    \hfill
    \begin{subfigure}[b]{0.4\textwidth}
        \centering
        \includegraphics[width=\textwidth]{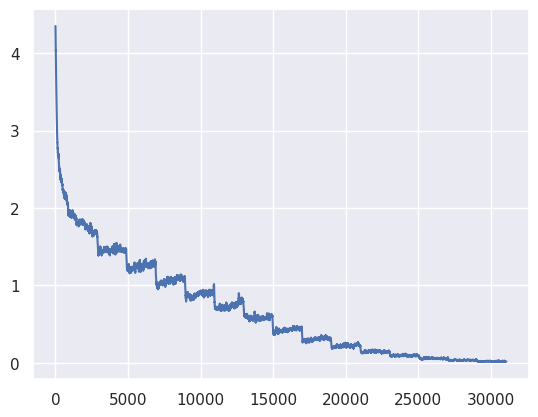}
        \caption{mT5 Bi-lingual eng-hi one-to-one}
        \label{fig:mT5-bi}
    \end{subfigure}
    \caption{Loss Convergence plots for Fine-tuning}
    \label{fig:mT5-bi-multi}
\end{figure}

For the fine-tuning task, the mT5 model demonstrated significant performance with a BLEU score of 14.1444 and a chrF score of 33.8278 for the bilingual translation task between English and Hindi. This indicates the model's strong capability in handling the en-hi translation pair. Additionally, we extended our experiments to include six different combinations involving English, Hindi, and Bengali for the multilingual mT5 model, exploring various language pairings to assess the model's robustness across different translation tasks.

In the case of the LLaMA 2 model, our evaluations revealed that the 1-1 configuration outperformed the 1-many model. This suggests that the one-to-one translation setup is more effective for maintaining high translation quality, outperforming the many-to-one setup in our experiments.

\begin{figure}
    \centering
    \includegraphics[scale=0.42]{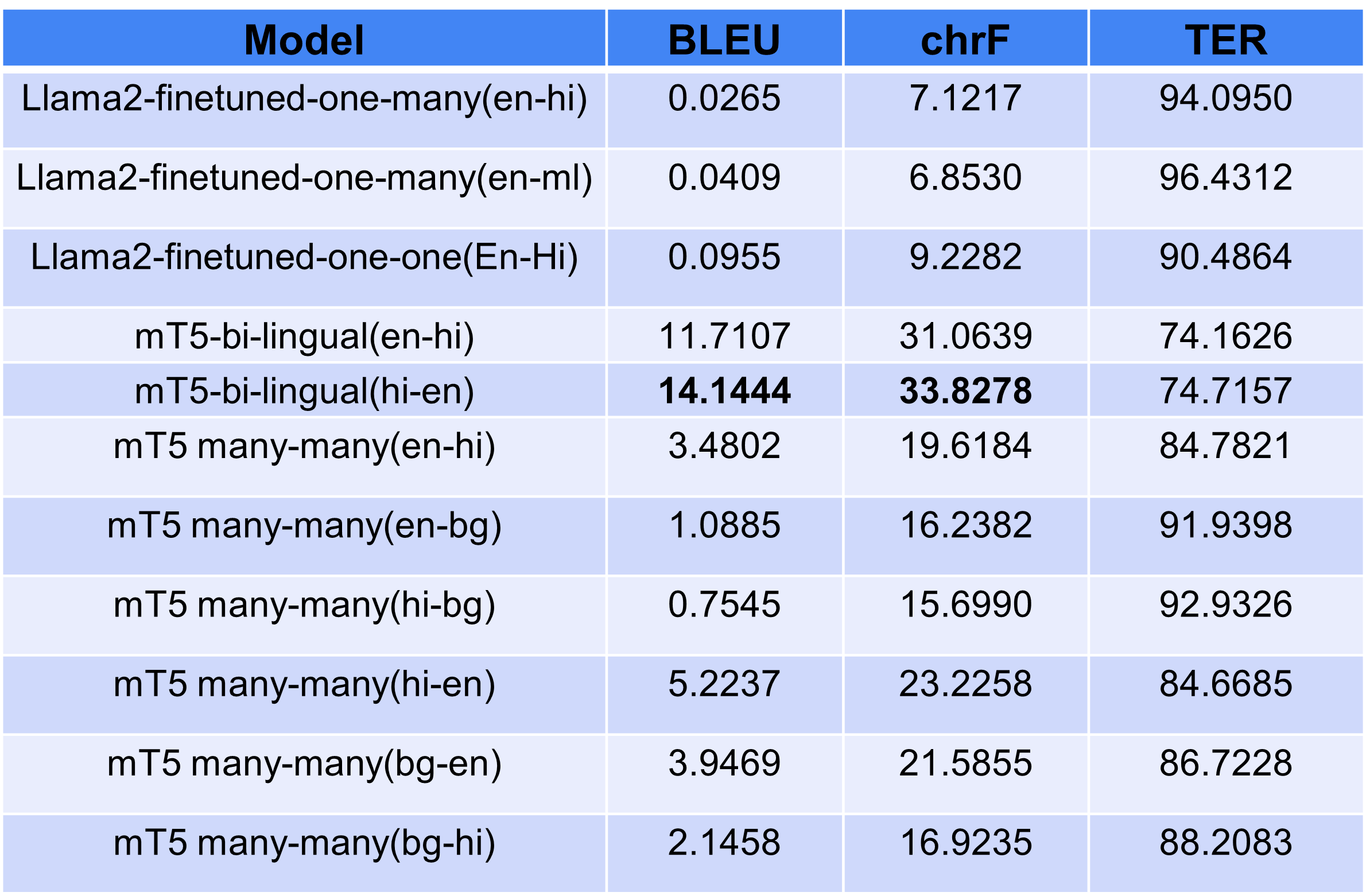}
    \caption{Results for Finetuning the models}
    \label{fig:results_finetune}
\end{figure}

\subsection{Baseline Model Development}
The baseline models for this study are built upon pre-trained models that have been trained on extensive datasets. Specifically, we utilize the mT5 model, which is pre-trained to understand multiple languages. This inherent multi-lingual capability of mT5 is leveraged in our fine-tuning process to adapt the model to the specific tasks at hand.

However, the problem statement for our study remains incomplete. Our primary objective is to compare the performance of Encoder-Decoder and Decoder-only models under similar training conditions. This comparison aims to evaluate the effectiveness of these models in multi-task learning scenarios, particularly in multi-lingual machine translation (MT). To ensure a comprehensive comparison, we will examine not only the overall performance metrics of the models but also their ability to handle different context lengths. Quantitative metrics will be employed to measure and interpret the models' performance with varying context lengths, providing deeper insights into their strengths and limitations. This thorough evaluation will help us understand which model architecture is better suited for multi-lingual and multi-task learning applications.

\begin{figure}
    \centering
    \includegraphics[scale=0.14=5]{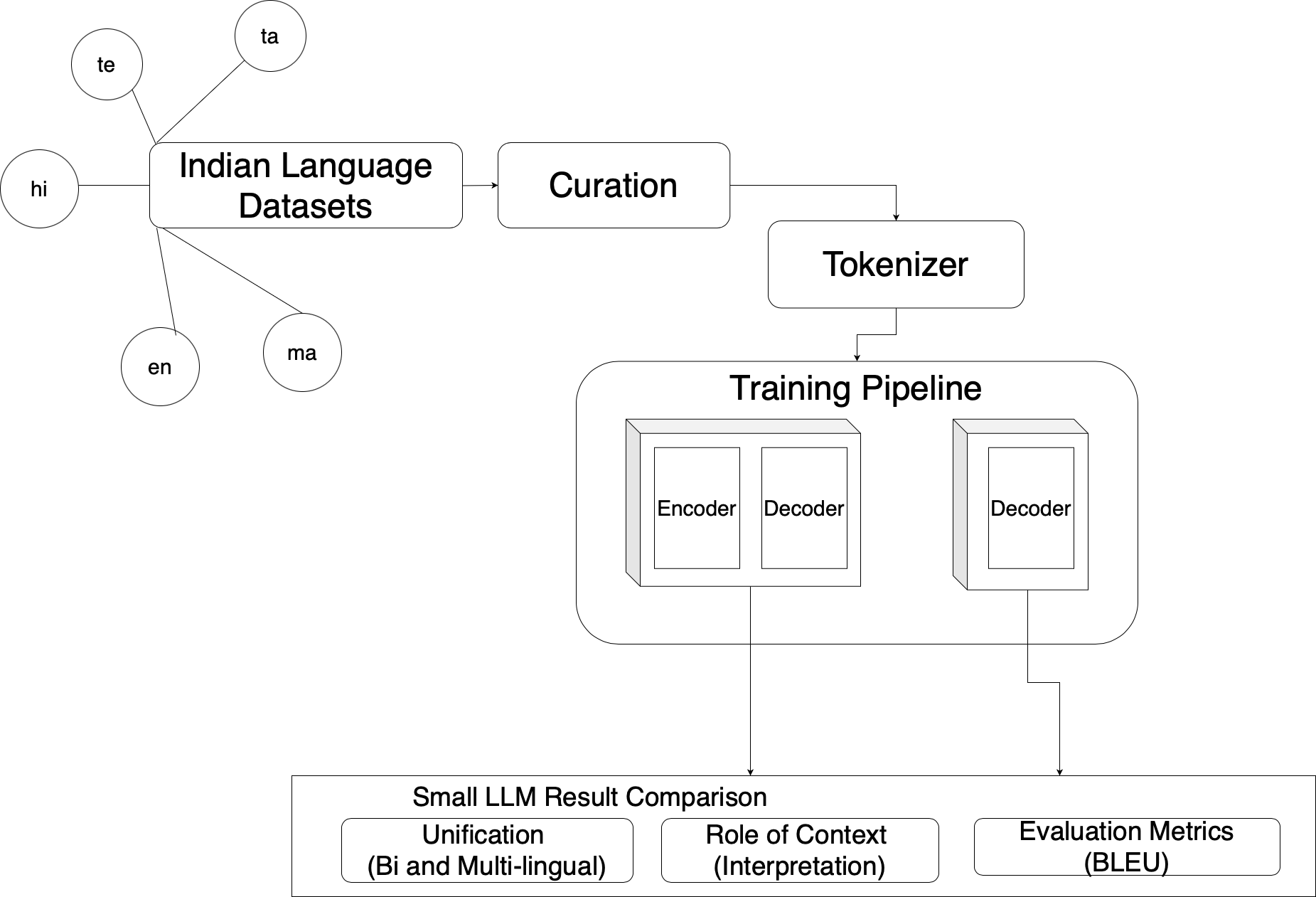}
    \caption{Proposed Methodology for Baseline Model Comparison}
    \label{fig:baseline-model}
\end{figure}

The proposed methodology of our multi-lingual Encoder-Decoder based architecture can be viewed in Figure \ref{fig:baseline-model}. We used the same datasets from the IndicTrans2  \cite{indictrans2} which is in-tune with our Indian baselines study. We go to the next step which is data-curation where we curate the dataset for better translation.

We wanted to write a model from scratch which we able to provide some construction, as using the pretrained model is more black boxed and less interpretable.
As creating and experimenting with the model comes out with their own challenges, we took some stable baseline models and we equated the parameters.
We use XLNet as a base model for implementing the MT task based learning as a Decoder-only model (Wu et al, 2021) and for Encoder-Decoder only model we use the IndicBART as a base model (Dabre et al, 2021) with the shared tokenizer.

\begin{table}[h!]
    \centering
    \begin{tabular}{|l|c|}
        \hline
        \textbf{Model Name} & \textbf{Trainable Parameters} \\ \hline
        XLNet Baseline      & 147,490,318                  \\ \hline
        Indic-BART Baseline & 145,339,392                  \\ \hline
    \end{tabular}
    \caption{Trainable parameters for XLNet and Indic-BART baseline models}
    \label{tab:model-parameters}
\end{table}

We utilized the datasets from IndicTrans2 \cite{indictrans2}, aligning with our focus on Indian language baselines. Following dataset selection, we curated the data to enhance translation quality. Our initial intention was to develop a model from scratch to ensure transparency and interpretability, as pre-trained models often function as black boxes. However, building and experimenting with custom models introduces significant challenges. To address these, we employed stable baseline models and ensured parameter equivalence for fair comparison. For the Decoder-only model, we chose XLNet as the base, implementing it for multi-task learning in machine translation, as demonstrated by \cite{wu2021low} For the Encoder-Decoder model, we used IndicBART as the foundation, based on the work of \cite{dabre2021indicbart}, Both models shared a common tokenizer to maintain consistency in data processing. This approach allowed us to systematically compare the performance and interpretability of Decoder-only and Encoder-Decoder models under similar conditions, providing insights into their respective strengths and weaknesses in multi-lingual and multi-task learning scenarios. As shown in Table \ref{tab:model-parameters}, the XLNet Baseline model has 147,490,318 trainable parameters, while the Indic-BART Baseline model has 145,339,392 parameters.

\subsubsection{Baseline Experimental Results}

\begin{figure}[ht]
    \centering
    \begin{subfigure}[b]{0.4\textwidth}
        \centering
        \includegraphics[width=\textwidth]{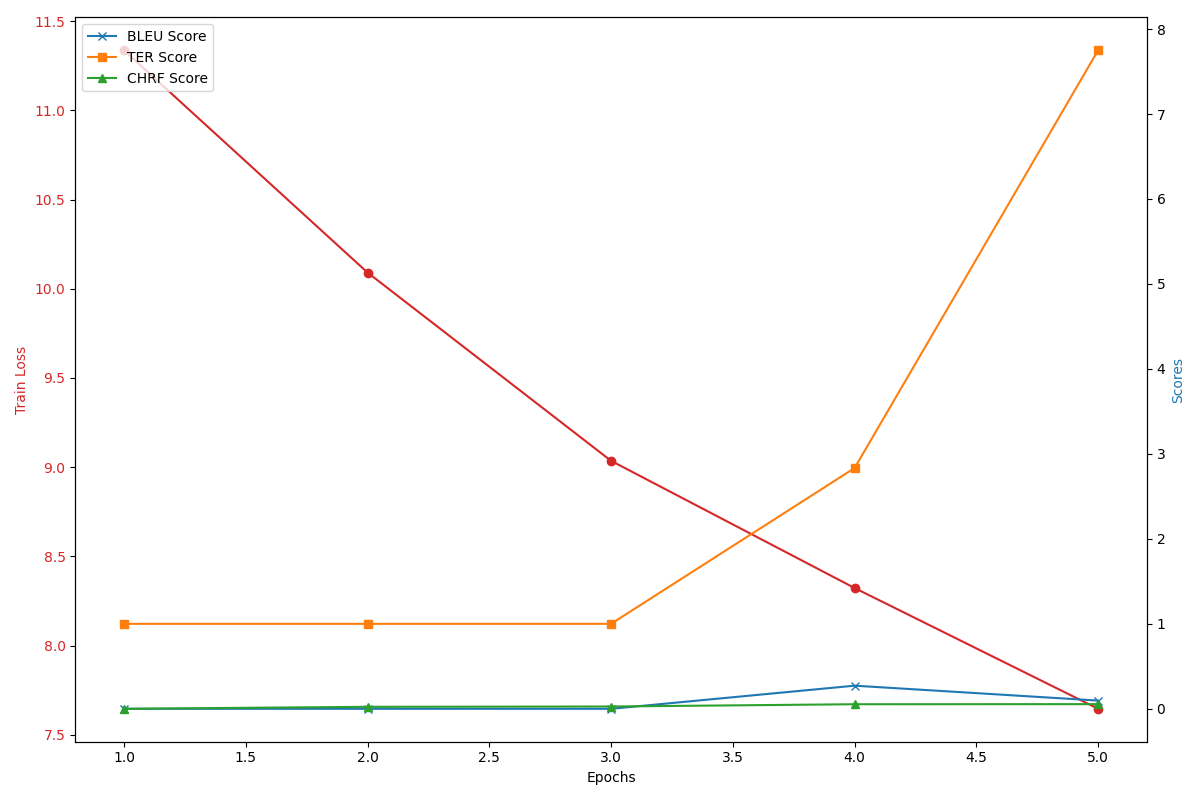}
        \caption{Encoder-Decoder One-to-One Eng to Hindi Results}
        \label{fig:image1}
    \end{subfigure}
    \hfill
    \begin{subfigure}[b]{0.4\textwidth}
        \centering
        \includegraphics[width=\textwidth]{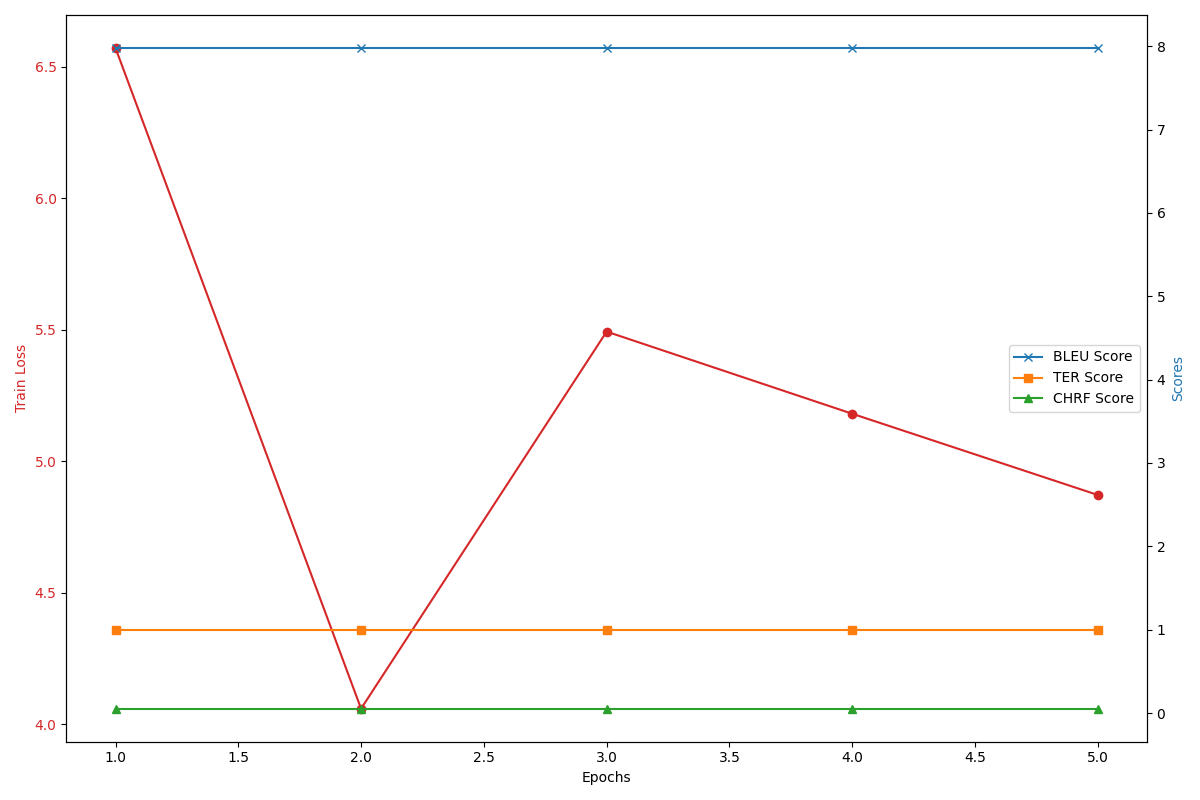}
        \caption{Decoder only One-to-One Eng to Hindi Results}
        \label{fig:image2}
    \end{subfigure}
    \caption{Loss Convergence, BLEU, chrF, TER}
    \label{fig:both_images}
\end{figure}

\begin{figure*}[ht]
    \centering
    \begin{subfigure}[b]{0.45\textwidth}
        \centering
        \includegraphics[width=\textwidth]{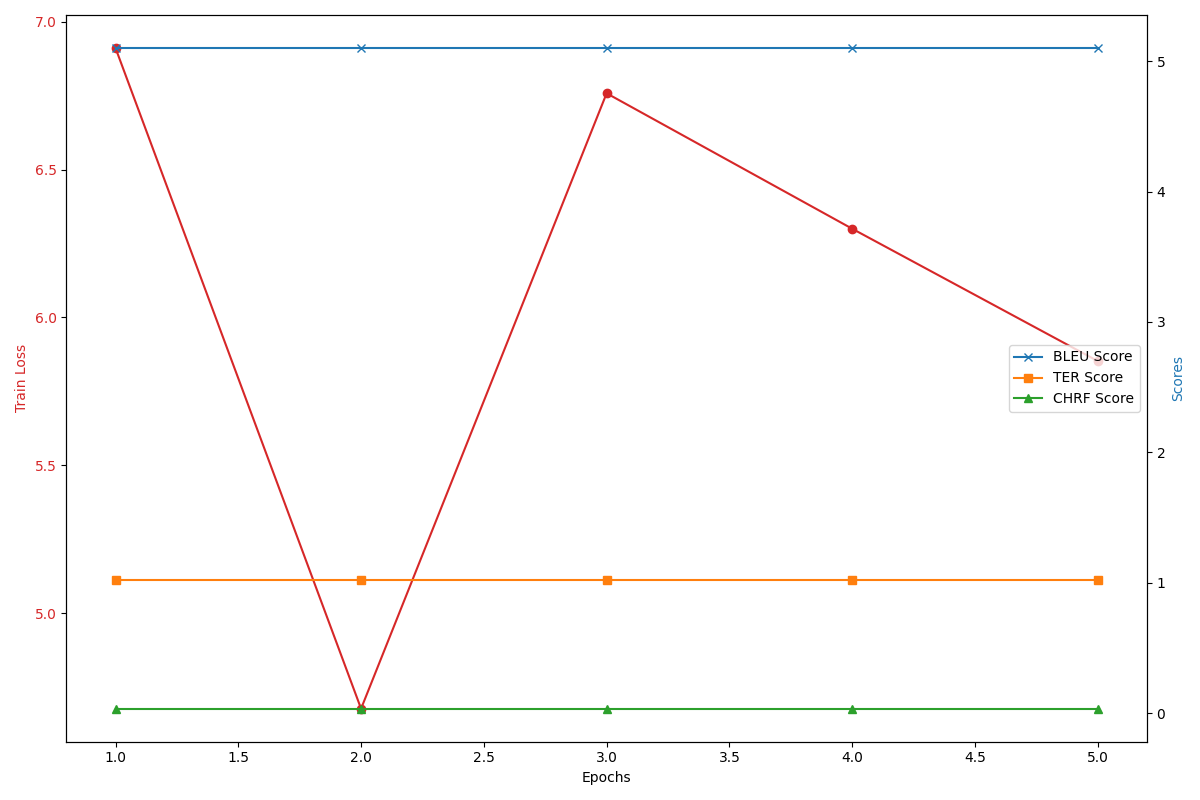}
        \caption{Decoder only One-to-Many English to (Hindi, Marathi) Translation}
        \label{fig:image1}
    \end{subfigure}
    \hfill
    \begin{subfigure}[b]{0.45\textwidth}
        \centering
        \includegraphics[width=\textwidth]{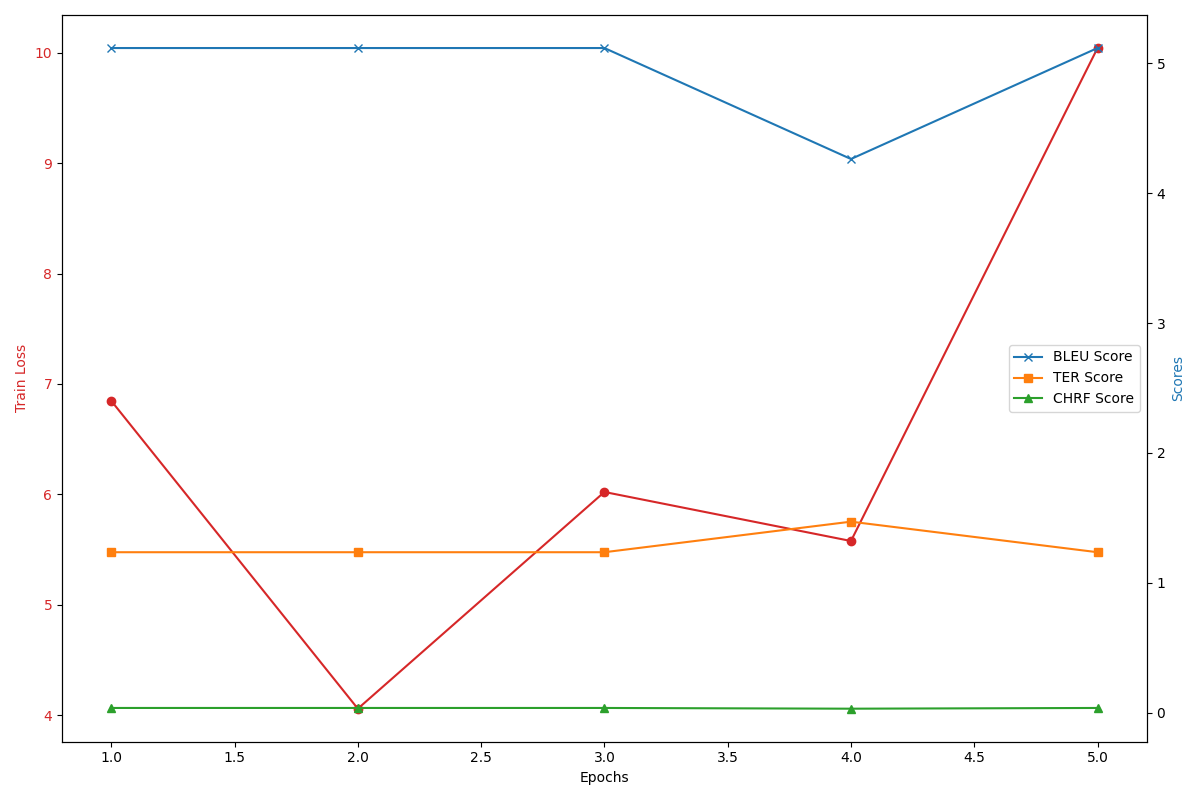}
        \caption{Decoder only Many-to-One (Hindi, Marathi) to English Translation}
        \label{fig:image2}
    \end{subfigure}

    \medskip

    \begin{subfigure}[b]{0.45\textwidth}
        \centering
        \includegraphics[width=\textwidth]{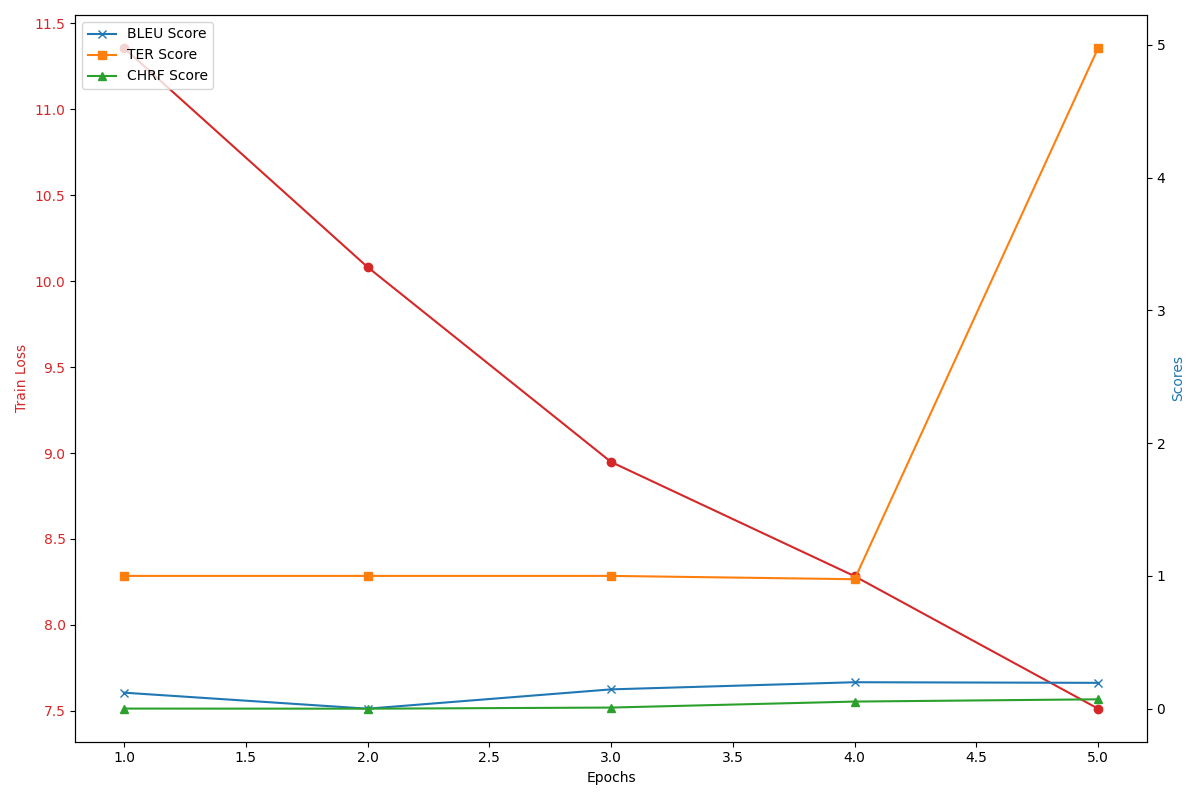}
        \caption{Encoder-Decoder One-to-Many English to (Hindi, Marathi) Translation}
        \label{fig:image3}
    \end{subfigure}
    \hfill
    \begin{subfigure}[b]{0.45\textwidth}
        \centering
        \includegraphics[width=\textwidth]{images/enc_dec_one2many.png}
        \caption{Many-to-One (Hindi, Marathi) to English Translation}
        \label{fig:image4}
    \end{subfigure}
    \caption{Loss Convergence, BLEU, chrF, TER}
    \label{fig:one2many}
\end{figure*}

In this section, we present the experimental results of our study, focusing on comparisons between one-to-one, one-to-many, many-to-one Encoder-Decoder, and Decoder-only models. Figure \ref{fig:one2many} illustrates the performance comparison between one-to-one and one-to-many Encoder-Decoder models. The BLEU scores for one-to-many model (Figure \ref{fig:one2many}(a)) indicate a slight improvement in translation quality over the one-to-one model (Figure \ref{fig:one2many}(b)), particularly in handling multiple outputs from a single input. Figure \ref{fig:both_images} presents the results for many-to-one and Decoder-only models. The many-to-one model (Figure \ref{fig:both_images}(a)) shows robust performance in aggregating multiple inputs into a single output, while the Decoder-only model (Figure \ref{fig:both_images}(b)) excels in generating fluent translations with reduced computational complexity.

\section{Conclusion and Future Work}

The Encoder-Decoder model has demonstrated reliable performance in our experiments, providing trustworthy results. However, the training paradigms for Decoder-only models differ significantly, as they are typically trained on next-word or next-character prediction tasks. The inherent differences in learning paradigms raise the question of how to achieve convergence in multilingual machine translation. Decoder-only models handle the starting positions of the source and target texts separately, posing unique challenges.

Additionally, the development of novel methods such as Streaming Self-Attention (SSA) marks a significant advancement. SSA enables the model to determine when it has sufficient context from the original text to begin translating accurately. This technique addresses some of the inherent challenges in translating long texts and could be crucial for improving the performance of Decoder-only models in multilingual settings. Future work should focus on refining these models and exploring ways to harmonize the learning paradigms of Encoder-Decoder and Decoder-only architectures. Further research is needed to fully exploit the potential of SSA and other innovative techniques to enhance translation accuracy and efficiency. This research will contribute to the broader goal of advancing machine translation technologies for multilingual applications.

\section*{Acknowledgements}
We extend our heartfelt gratitude to the IIITH Administration for providing excellent resources and amenities, which have been instrumental in the successful completion of this work. We are deeply appreciative of the guidance and support from the IASNLP Coordinators, Dr. Parameswari and Dr. Rahul Mishra, whose expertise and encouragement have greatly contributed to our research. Our sincere thanks go to our mentor, Mr. Yash Bhaskar, for his invaluable advice and mentorship throughout this project. We would also like to acknowledge the contributions of our pre-school speakers: Prashanth Kodali, Aparajitha, Priyanka Dasari, Aadya Ranjan, and Sankalp Bahad, whose insights and feedback have been immensely helpful.

\end{document}